\documentclass[sigconf,natbib]{acmart}
\usepackage{multirow}
\usepackage{subcaption}
\AtBeginDocument{%
  }

\setcopyright{acmlicensed}
\copyrightyear{2025}
\acmYear{2025}
\acmDOI{10.1145/XXXXXXX.XXXXXXX} % Replace with the correct DOI

\acmConference[CIKM '25]{Proceedings of the 34th ACM International Conference on Information and Knowledge Management}{November 10--14, 2025}{Seoul, Republic of Korea}

\acmBooktitle{Proceedings of the 34th ACM International Conference on Information and Knowledge Management,  2025, Seoul, Republic of Korea}
\acmISBN{978-1-4503-XXXX-X/18/06}

\begin{document}

\title[A Comparative Study of Multimodal, Caption-Based, and Hybrid Slide Retrieval Techniques]{What’s the Best Way to Retrieve Slides? A Comparative Study of Multimodal, Caption-Based, and Hybrid Retrieval Techniques}

\author{Petros Stylianos Giouroukis}
\email{pgiouroukis@gmail.com}
\orcid{}
\affiliation{%
  \institution{Aristotle University of Thessaloniki}
  \city{Thessaloniki}
  \country{Greece}
}

\author{Dimitris Dimitriadis}
\email{dndimitri@csd.auth.gr}
\orcid{}
\affiliation{%
  \institution{Aristotle University of Thessaloniki}
  \city{Thessaloniki}
  \country{Greece}
}

\author{Dimitrios Papadopoulos}
\email{dpapado1@its.jnj.com}
\orcid{}
\affiliation{%
  \institution{Johnson \& Johnson}
  %\city{Thessaloniki}
  \country{United States}
}

\author{Zhenwen Shao}
\email{zshao5@its.jnj.com}
\orcid{}
\affiliation{%
  \institution{Johnson \& Johnson}
  %\city{Thessaloniki}
  \country{United States}
}

\author{Grigorios Tsoumakas}
\email{greg@csd.auth.gr}
\orcid{0000-0002-7879-669X}
\affiliation{%
  \institution{Aristotle University of Thessaloniki}
  \city{Thessaloniki}
  \country{Greece}
}

\renewcommand{\shortauthors}{P.S. Giouroukis, D. Dimitriadis, D. Papadopoulos, Z. Shao, and G. Tsoumakas}

%%
%% The abstract is a short summary of the work to be presented in the article.
\begin{abstract}
Slide decks, serving as digital reports that bridge the gap between presentation slides and written documents, are a prevalent medium for conveying information in both academic and corporate settings. Their multimodal nature, combining text, images, and charts, presents challenges for retrieval-augmented generation systems, where the quality of retrieval directly impacts downstream performance. Traditional approaches to slide retrieval often involve separate indexing of modalities, which can increase complexity and lose contextual information. This paper investigates various methodologies for effective slide retrieval, including visual late-interaction embedding models like ColPali, the use of visual rerankers, and hybrid retrieval techniques that combine dense retrieval with BM25, further enhanced by textual rerankers and fusion methods like Reciprocal Rank Fusion. A novel Vision-Language Models-based captioning pipeline is also evaluated, demonstrating significantly reduced embedding storage requirements compared to visual late-interaction techniques, alongside comparable retrieval performance. Our analysis extends to the practical aspects of these methods, evaluating their runtime performance and storage demands alongside retrieval efficacy, thus offering practical guidance for the selection and development of efficient and robust slide retrieval systems for real-world applications.
\end{abstract}

%\begin{CCSXML}




\maketitle

\section{Introduction}

% Slides combine text, images, charts, and multimedia elements to convey information in an engaging and digestible way. The standard need that slides serve for academics is the visual support of talks delivered at conferences and workshops, as well as lectures given in classrooms or online. In the modern day business environment however, slides are often used as standalone document meant for independent reading, and {\em slide deck} is used to differentiate this role of slides from standard {\em presentation slides} \cite{smith_2021}. Presentation slides prioritize simplicity, with minimal text, impactful visuals, and a design that emphasizes audience focus on the presenter. Slide decks, on the other hand, often contain detailed information, comprehensive data, and fully-formed arguments. Slide decks can therefore be defined as digital reports that stand in the middle between presentation slides and written reports. They contain more text and less images than a standard presentation, but less text and more images than a typical written report. 
Slides combine text, images, charts, and multimedia elements to convey information effectively. While academics use slides for visual support during conferences and lectures, businesses increasingly employ them as standalone documents for independent reading. This distinction separates {\em slide decks} from standard {\em presentation slides} \cite{smith_2021}. Presentation slides emphasize simplicity with minimal text and impactful visuals designed to focus audience attention on the presenter. Slide decks contain detailed information, comprehensive data, and fully-formed arguments, functioning as digital reports positioned between presentation slides and written documents. They feature more text than standard presentations but more visuals than typical written reports.

% The high volume of slide decks, particularly in large organizations, contains a wealth of information and knowledge, making them ideal sources for corporate retrieval augmented generation (RAG)~\cite{10.5555/3495724.3496517} systems. The quality of a RAG system can only be as high as the quality of its retrieval component, as the large language model (LLM) component cannot respond properly to user queries concerning enterprise knowledge if it doesn't have the right information in its context. This motivates us to investigate the ideal architecture for a text-to-slide retrieval component.  
Large organizations' extensive slide deck repositories contain valuable knowledge, making them ideal sources for corporate retrieval augmented generation (RAG) \cite{10.5555/3495724.3496517} systems. RAG system quality depends directly on retrieval quality, since LLMs cannot properly address enterprise queries without relevant contextual information. This motivates us to investigate optimal text-to-slide retrieval architectures.

% A common way to deal with the multimodality of slide decks is to index separately the different modalities. Optical character recognition (OCR) technology can be applied to extract the textual content of the slides, which can subsequently be directly indexed using traditional sparse models, such as BM25~\cite{Robertson2009ThePR} or encoded in dense vector representations \cite{karpukhin-etal-2020-dense}. Figures and tables can be extracted by object detection models trained on relevant datasets~\cite{Pfitzmann2022}, and then encoded by visual and table understanding~\cite{Nassar2022} models respectively. 
Traditionally, slide deck multimodality is handled by separate indexing.  Optical Character Recognition (OCR) extracts text for sparse (BM25~\cite{Robertson2009ThePR}) or dense~\cite{karpukhin-etal-2020-dense} indexing. Figures and tables, identified by object detection~\cite{Pfitzmann2022}, are encoded via visual and table understanding models~\cite{Nassar2022} respectively.

% Working separately with each modality increases the complexity of the retrieval component and can lead to  loss of information deriving from their coexistence in the same slide~\cite{ma-etal-2024-unifying}. To deal with these issues, Document Screenshot Embedding (DSE)~\cite{ma-etal-2024-unifying} proposed using a large vision-language model (VLM) to directly encode entire slide screenshots into dense vector embeddings. Other strategies transform the multimodal slide retrieval problem into a textual one by first employing VLMs to generate descriptive captions of the slides. This allows the leveraging of well-established textual retrieval infrastructure, including techniques like hybrid search combining sparse (e.g., BM25) and dense retrievers, followed by sophisticated rerankers to refine the results. More recently, models like ColPali have introduced a different paradigm by combining VLMs with the late interaction mechanism inspired by ColBERT.
Working separately with each modality complicates the retrieval and can lose co-occurrence information~\cite{ma-etal-2024-unifying}. Document Screenshot Embedding (DSE)~\cite{ma-etal-2024-unifying} addresses this using a large Vision-Language Model() to directly encode entire slide screenshots. Other strategies employ VLMs for slide captioning, transforming multimodal retrieval into a textual problem. This leverages established textual infrastructure, including hybrid search (combining sparse and dense retrievers) and rerankers. More recently, ColPali \cite{faysse2025colpaliefficientdocumentretrieval} introduced a different paradigm, combining VLMs with the ColBERT-inspired \cite{10.1145/3397271.3401075} late interaction mechanism.

For RAG systems, balancing accuracy and response time is essential, particularly for real-world chatbot applications. While many advanced retrieval solutions maximize accuracy, their longer processing times make them unsuitable for industry chatbot RAGs requiring real-time responses. Systems emphasizing both speed and accuracy better meet the strict response time requirements of interactive chatbot interfaces. Additionally, smaller embedding sizes significantly reduce storage requirements, lowering infrastructure and maintenance costs for organizations. While academia typically prioritizes peak accuracy metrics, industry applications must balance multiple quality of dimensions to deliver functional user experiences.

% This paper presents a comparative study of various retrieval methodologies specifically for slide decks. We analyze their retrieval effectiveness alongside practical considerations crucial for real-world deployment, such as storage demands and retrieval latency, offering insights into the trade-offs involved in selecting an optimal slide retrieval system under varying resource constraints.
This paper compares slide deck retrieval methodologies, analyzing retrieval effectiveness alongside practical deployment considerations like storage demands and latency, providing insights into trade-offs for selecting optimal systems under varying resource constraints. In detail, the main contributions of this paper are:
\begin{itemize}
\item A comparative analysis guiding the practical design of real-world slide retrieval pipelines.
\item VLM-based captioning evaluated as competitive in retrieval performance, storage, and inference time, approaching state-of-the-art (SOTA) results.
\item New SOTA results for slide retrieval achieved through an engineered, high-performing pipeline.
\end{itemize}
This work is structured as follows: Section 2 reviews related work. Section 3 details the experimental setup, including datasets, retrieval methods, captioning models, rerankers, baselines, evaluation metrics, and infrastructure. Section 4 presents and discusses the results of our comparative study. Finally, Section 5 concludes the paper and outlines future research directions.

\section{Related Work}

% We organize the related work into four sections. First, we examine VLMs, as our method utilizes such a model for image captioning. Next, we discuss works related to multimodal document retrieval. Third, we review available slide datasets. Finally, we mention approaches that leverage two-stage retrieval, as our work adopts such approaches. 

\subsection{Vision-Language Models}

VLMs have revolutionized multimodal understanding by enabling joint visual and textual processing. Early foundational work like CLIP~\cite{radford2021learning} established visual-textual alignment through contrastive learning, while BLIP~\cite{pmlr-v162-li22n} advanced pre-training frameworks for both understanding and generation tasks. Recent state-of-the-art models including GPT-4 \cite{openai2024gpt4technicalreport} and Gemini-1.5 \cite{geminiteam2024gemini15unlockingmultimodal} demonstrate strong performance across diverse visual tasks from object recognition to image captioning. A key advancement has been developing more efficient architectures where smaller models like PaliGemma 2 \cite{steiner2024paligemma2familyversatile} and Qwen2-VL \cite{Qwen2-VL} achieve competitive results, particularly with fine-tuning, making them more practical for deployment. This efficiency is complemented by improved pre-training strategies such as visual instruction tuning in LLaVA \cite{liu2023visualinstructiontuning} and Phi-3.5-Vision \cite{abdin2024phi3technicalreporthighly}, enhancing capabilities across visual understanding tasks.

% Recent VLMs such as Molmo-7B \cite{deitke2024molmopixmoopenweights} and Moondream2\footnote{\url{https://huggingface.co/vikhyatk/moondream2}} have further enhanced their capabilities by training on diverse datasets that encompass OCR, image captioning, as well as chart and table comprehension. Similarly, the Gemma 3 family of models use techniques like knowledge distillation and Quantization-Aware Training (QAT) to offer strong performance with significantly reduced memory footprints, making them highly suitable for deployment within constrained compute budgets. Our work leverages these models' zero-shot capabilities to generate comprehensive captions for slides without requiring any domain-specific fine-tuning.
Recent VLMs like Molmo-7B \cite{deitke2024molmopixmoopenweights} and Moondream2\footnote{\url{https://huggingface.co/vikhyatk/moondream2}} are trained on diverse datasets including OCR, image captioning, and chart/tab\-le comprehension, enhancing their capabilities. Similarly, the Gem\-ma3 \cite{gemmateam2025gemma3technicalreport} family of models employs techniques like knowledge distil\-lation and Quantization-Aware Training for strong performance with reduced memory footprints, suiting constrained compute bu\-dgets. Our work leverages these models' zero-shot capabilities for comprehensive slide captioning, eliminating the need for doma\-in-specific fine-tuning.

\subsection{Multimodal Document Retrieval}

DSE~\cite{ma-etal-2024-unifying} employs a bi-encoder architecture with the Phi-3-vision VLM to directly encode document screenshots into vector embeddings. It unifies visual context, layout, and textual details, bypassing OCR or content extraction. Optimized via contrastive learning, DSE shows superior performance on text-intensive mixed-modality documents and is evaluated here as a direct slide retrieval technique.

ColPali~\cite{faysse2025colpaliefficientdocumentretrieval} adapts ColBERT's~\cite{10.1145/3397271.3401075} late-interaction for multimodal retrieval. Images are encoded into multi-vector representations via patch-level visual embeddings, and queries are similarly processed per-token. A late-interaction mechanism then compares query and visual tokens, capturing finer semantic details than single-vector models.

% The combination of multimodal retrieval with RAG in an industrial scenario involving a very small, and closed source, collection of 20 PDFs (manuals and software documentation for devices) is explored in \cite{riedler2024textoptimizingragmultimodal}. Like most approaches besides DSE, they extracted text and images as separate entities and text was further chunked to support the RAG workflow. Interestingly, they found that textual retrieval on top of captioning the images leads to improved RAG metrics compared to separate indexing of images and text. While their findings support the potential of caption-based retrieval in a specific RAG pipeline with pre-extracted images, our research takes this concept further. We also investigate a captioning pathway but apply it directly to whole slides as single units, thereby avoiding preliminary object extraction or modality separation.
% Riedler et al. \cite{riedler2024textoptimizingragmultimodal} explored multimodal retrieval with RAG on a small, closed collection of 20 PDFs, extracting text and images separately. They found that textual retrieval on image captions improved RAG metrics compared to separate indexing. While their findings support caption-based retrieval for pre-extracted images, our research extends this by applying captioning directly to whole slides, thus avoiding preliminary object extraction or modality separation.
Riedler et al. \cite{riedler2024textoptimizingragmultimodal} showed textual retrieval on image captions improved RAG for a small PDF collection with pre-extracted modalities. While their work supports captioning for extracted images, our research applies it directly to whole slides, avoiding such preliminary separation.

\begin{figure*}[!ht]
    \centering
    \includegraphics[width=0.99\textwidth]{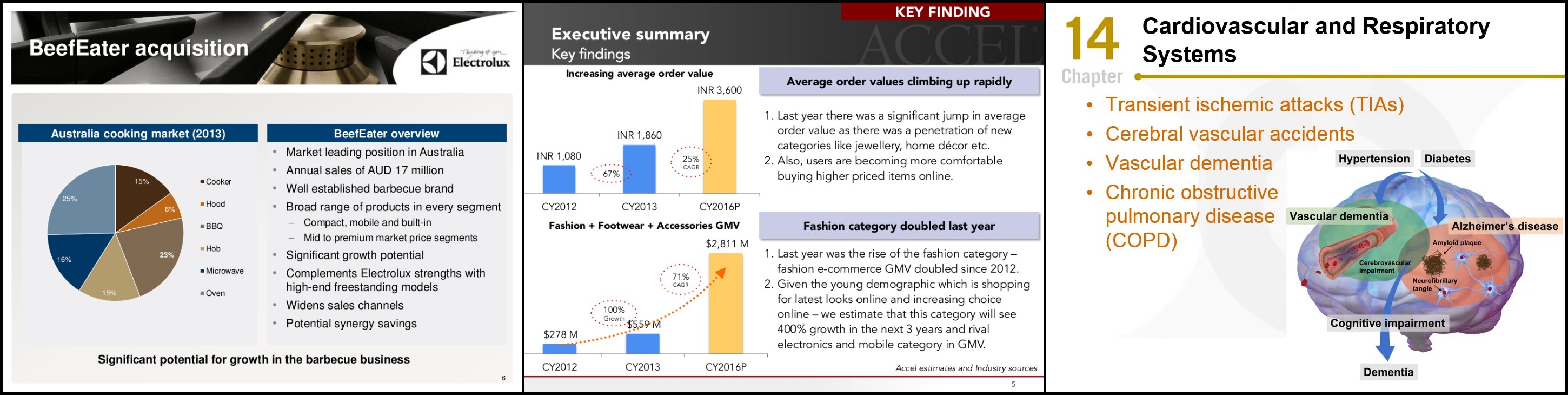}
    \caption{
    % \textbf{Example slides, queries, and answers from the SlideVQA and LPM datasets.}
    Examples of slides from the SlideVQA and LPM datasets. From left to right: First two slides show examples from SlideVQA, with queries `In Australia cooking market how much does cooker contain?' and `How much difference in INR is there between the average order value of CY2013 and that of CY2012?' respectively. The last slide is from the LPM dataset, where queries are derived from lecture transcripts. Part of this slide's transcript includes `People tend to think that if it's dementia that it must be a brain-related problem...'.
    }
    \Description{
    Four examples from the SlideVQA and LPM datasets. Each example includes a slide image on the left, a query below the slide, and the corresponding answer below the query. The top two examples are from SlideVQA, and the bottom two are from LPM. 
    }
    \label{fig:dataset_examples_combined}
\end{figure*}

\subsection{Slide Datasets}
%others to consider adding in the future 
%DreamStruct: Understanding Slides and User Interfacs via Synthetic Data Generation

Several datasets containing slides come from the domain of academic lecturing~\cite{nguyen2014,lectureVideoDB2018,Li2019,lee2023lecture,chen-etal-2024-m3av}. However none of these contain pairs of queries and reference slides, as they were created to support different tasks, including lecture retrieval~\cite{nguyen2014}, localizing and recognizing text~\cite{lectureVideoDB2018}, lecture preparation and reading list generation~\cite{Li2019}, slide and script generation, automatic speech recognition and text-to-speech synthesis~\cite{chen-etal-2024-m3av}.

For evaluating slide retrieval systems, two datasets stand out by providing question-answer pairs that can serve as relevance judgments. SlideVQA \cite{tanaka2023slidevqa} is a dataset of 2,619 slide decks paired with visual question-answer queries, emphasizing the retrieval of relevant slides to answer given queries. Lecture Presentations Multimodal (LPM) \cite{lee2023lecture} is a collection of lecture slides with corresponding multimodal data (e.g., figures, transcribed lecture speech), suitable for exploring slide retrieval in an academic setting. We discuss both datasets in more detail in Section \ref{sec:exp_setup}.

\subsection{Two-stage Retrieval}

Information retrieval methods are broadly categorized into sparse and dense retrievers \cite{li2022interpolate}. A common two-stage approach first uses sparse techniques to retrieve initial documents, then applies dense methods for reranking using more complex semantic understanding via embeddings or deep learning models. Many studies have utilized this framework ~\cite{almeida2023bit,almeida2024bit,lesavourey2024enhancing,cserbetcci2024hu, lesavourey2023bioasq}, traditionally with textual rerankers, though the field has advanced to include multimodal rerankers such as \texttt{jina-reranker-m0}\footnote{\url{https://huggingface.co/jinaai/jina-reranker-m0}} that refine results based on image content. Alternatively, hybrid fusion methods combine outputs from multiple retrievers to leverage their unique relevance characteristics \cite{bruch2023analysis, li2022interpolate}, often followed by reranking for further performance enhancement. This paper evaluates such hybrid approaches, including sparse and dense retriever combinations with reranking, alongside other slide retrieval methodologies.

\section{Experimental Setup}
\label{sec:exp_setup}

% This section outlines the experimental configurations used in our experiments. We begin by presenting the various models and tools employed. Next, we describe the two datasets utilized, followed by a discussion of the baselines, other approaches, and our ablation study setup. We conclude with an overview of the evaluation process.

\subsection{Datasets}

\subsubsection{SlideVQA}

SlideVQA \cite{tanaka2023slidevqa} is a dataset developed to advance VQA research on presentation slides, consisting of 2,619 slide decks from SlideShare\footnote{\href{https://www.slideshare.net}{https://www.slideshare.net}} comprising over 52,000 slides and 14,484 VQA samples. The dataset evaluates models' ability to retrieve relevant content and reason across textual and visual slide elements, with each sample associated with ground truth slides for evaluation and including both single-hop questions (answerable with one slide) and multi-hop questions (requiring multiple slides). Following \cite{ma-etal-2024-unifying}, we modify the original within-deck retrieval task to an open-domain setting, requiring retrieval of top-kk
k slides from an entire pool spanning multiple decks rather than limiting search to 20 slides within the same deck. This modification better simulates real-world industry scenarios where businesses maintain repositories of multiple slide decks and queries are submitted without prior knowledge of which deck contains relevant information.

\subsubsection{LPM}

The LPM dataset \cite{lee2023lecture} is a resource for training models on multimodal lecture slide content, containing 9,031 slides with corresponding spoken text, visual elements, and OCR-extracted text from 334 educational videos (187 hours) spanning 35 courses across biology, anatomy, psychology, dentistry, public speaking, and machine learning. The dataset includes 8,598 visual figures comprising natural images (45.1\%), diagrams (46.7\%), tables (3.5\%), and equations (4.6\%), with slides averaging 26 words and 0.94 images each. Developed from curated YouTube lecture videos, the dataset employs Google ASR for speech transcription and Tesseract OCR for slide text extraction, as well as manual annotations for slide segmentation, figure bounding boxes, and visual element labeling.

\subsection{Visual Retrieval Methods}

We evaluate direct multimodal retrieval approaches that process slide images without intermediate modality conversion. One such method is DSE \cite{ma-etal-2024-unifying}, for which we include both fine-tuned and zero-shot performance scores as reported by the authors for SlideVQA, and execute the publicly available model\footnote{\url{https://huggingface.co/Tevatron/dse-phi3-v1.0}} in zero-shot setting for LPM. Another visual retrieval paradigm is ColPali\footnote{\url{https://huggingface.co/vidore/colpali-v1.3}} \cite{faysse2025colpaliefficientdocumentretrieval}, which adapts the late-interaction mechanism from ColBERT to the multimodal domain. Slide images are tokenized into 1031 patch-level visual embeddings, resulting in a multi-vector representation of shape $1031 \times 128$, while textual queries are similarly processed into per-token embeddings of dimension 128.

\begin{figure}[h!]
    % \small
    \centering
    \begin{subfigure}[b]{0.45\textwidth}
        \includegraphics[width=\textwidth]{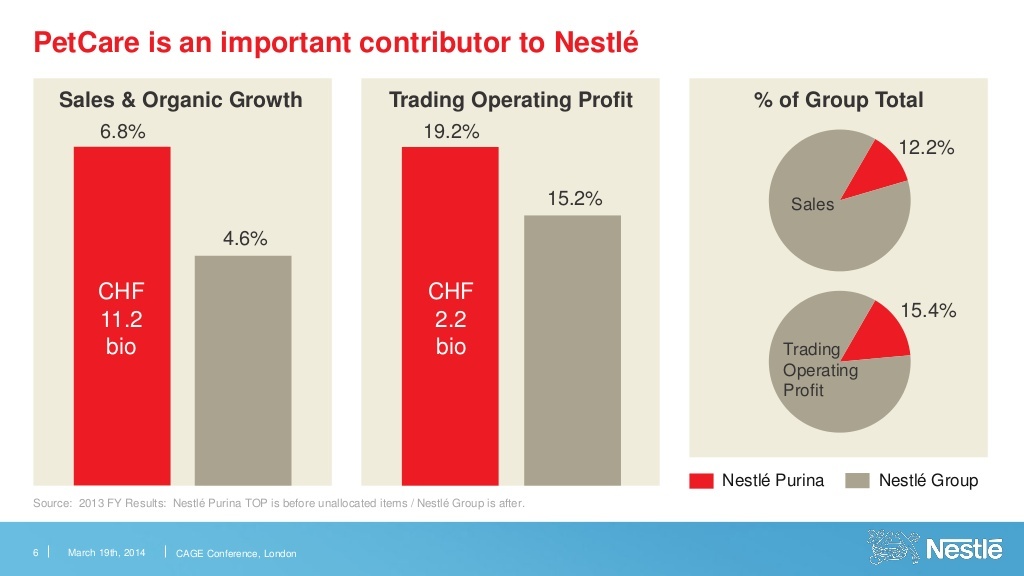}
        % \caption{Input slide.}
        \label{fig:input_slide}
    \end{subfigure}
    \hfill
    \begin{subfigure}[b]{0.45\textwidth}
        \fbox{\parbox{\textwidth}{
            \textbf{Molmo-7B-D-0924:} This slide presents financial data for Nestlé’s PetCare division. The title at the top states “PetCare is an important contributor to Nestlé.” The slide contains three main sections: 1. Sales \& Organic Growth: showing 6.8\% for Nestlé Purina and 4.6\% for Nestlé Group. 2. Trading Operating Profit: displaying 19.2\% for Nestlé Purina and 15.2\% for Nestlé Group. 3. \% of Group Total: illustrating 12.2\% for Sales and 15.4\% for Trading Operating Profit. The data indicates strong performance for Nestlé Purina in both sales growth and operating profit. The Nestlé Group’s figures are slightly lower in both categories. The slide includes a source note mentioning “2013 FY Results” and references a CAGE Conference in London on March 19, 2014.
        }}
        % \caption{Caption generated by Molmo-7B-D-0924.}
        \label{fig:molmo_caption}
    \end{subfigure}
    % \\
    % \begin{subfigure}[b]{0.45\textwidth}
    %     \fbox{\parbox{\textwidth}{
    %         \textbf{Moondream2:} The slide is titled "PetCare is an important contributor to Nestlé" and presents three key metrics related to Nestlé's performance. The first bar graph compares sales and organic growth, with CHF representing CHF sales and bio representing organic growth. The second bar graph shows the trading operating profit, with CHF representing CHF sales and bio representing organic growth. The third bar graph displays the percentage of group total, with CHF representing CHF sales and bio representing organic growth. The bottom right corner of the slide includes a legend indicating that Nestlé Purina is red and Nestlé Group is grey. 
    %     }}
        % \caption{Caption generated by Moondream2.}
        % \label{fig:moondream_caption}
    % \end{subfigure}
    
    \caption{Example of a SlideVQA slide and the corresponding caption generated by the Molmo-7B-D-0924 model.}
    \label{fig:caption_comparison}
    \Description{An image of a financial data slide from the SlideVQA dataset is on the top. On the bottom, its corresponding textual caption generated by the Molmo-7B-D-0924 model is shown within a bordered box.}
\end{figure}

\subsection{Captioning-based Retrieval}

% Our study emphasizes caption-based retrieval due to preliminary experiments indicating that it consistently outperformed purely visual retrieval methods, unless these methods explicitly incorporated textual and structural understanding, such as DSE and ColPali. A significant portion of our investigation focuses on transforming the multimodal slide retrieval problem into a textual one by first generating descriptive captions for each slide. This approach allows us to leverage established and powerful text-based retrieval techniques.

Our focus on captioning-based retrieval stems from initial experiments indicating its superior performance over many conventional visual retrieval methods, particularly those not explicitly optimized for extracting and understanding textual and structural information from slides. Approaches like DSE and ColPali, which do account for such multimodal intricacies, were found to be competitive. The captioning approach also allows us to leverage established and powerful text-based retrieval techniques.

% In our initial exploratory phase for slide-to-caption conversion, we experimented with Moondream2, a 2-billion parameter model, and Molmo-7B-D-0924 \cite{deitke2024molmopixmoopenweights}, a 7-billion parameter model. Preliminary qualitative evaluation revealed that captions generated by Molmo-7B-D-0924 were superior in quality, providing more detailed and accurate descriptions of slide content. Additionally, Molmo demonstrated enhanced OCR capabilities compared to Moondream2 Based on these findings, Molmo-7B-D-0924 was selected as one of the primary models for generating captions for the slides in all subsequent experiments. To further assess the impact of caption quality and style from a more powerful model, we also employed Gemma3-27B-IT. This is an instruction-tuned version of Google's Gemma 3 model with 27 billion parameters. 
In our initial slide-to-caption experiments, we compared Moondream2 \footnote{\url{https://huggingface.co/vikhyatk/moondream2}}(2B parameters),  Molmo-7B-D-0924 \cite{deitke2024molmo} \footnote{{\url{https://huggingface.co/allenai/Molmo-7B-D-0924}}} (7B parameters) and Gemma3-27B-IT-QAT\footnote{\url{https://huggingface.co/google/gemma-3-27b-it-qat-q4_0-gguf}}. Qualitative evaluation showed Molmo and Gemma3 produced superior captions with more detail, accuracy, and better OCR capabilities. Consequently, they were selected for further investigation.

Both Molmo-7B-D-0924 and Gemma3-27B-IT were used in a zero-shot manner. To guide caption generation, we used the instruction prompt: ``{\em“This is a presentation slide. Provide a detailed caption that will be used in a RAG pipeline. If you see any charts, tables, diagrams etc, make sure to explain what you see. Don't provide any additional information or explanations e.g. about colors and backgrounds. Start doing the captioning immediately.”}''. An example of an input slide alongside the detailed caption generated by Molmo-7B-D-0924 using this prompt is presented in Fig. ~\ref{fig:caption_comparison}.

% Once these captions are generated and indexed, they form the basis for several textual retrieval approaches investigated in this study. Beyond standard BM25, which is applied directly to these captions, we explore two other dense representation strategies. First, for standard dense retrieval, the generated slide captions are encoded using the NV-Embed-2 (8B parameters) identified as the top performer on the Hugging Face MTEB Leaderboard at the time of experimentation. The dimensionality of the embedding vectors produced by this model is 4,096. Textual queries are encoded using the same model for performing semantic similarity searches against the caption embeddings.
The generated captions form the basis for several textual retrieval approaches. We evaluate BM25 applied directly to these captions alongside two dense representation strategies. For standard dense retrieval, slide captions are encoded using NV-Embed-2 \cite{lee2025nvembedimprovedtechniquestraining} \footnote{\url{https://huggingface.co/nvidia/NV-Embed-v2}} (8B parameters), the top performer on the Hugging Face MTEB Leaderboard \footnote{\url{https://huggingface.co/spaces/mteb/leaderboard}} at experimentation time, producing 4,096-dimensional embeddings. Textual queries use the same model for semantic similarity searches against caption embeddings.

% Second, to explore the potential of late-interaction mechanisms for textual retrieval on captions, we adapted the ColPali architecture in an unconventional manner, termed "Textual ColPali". The motivation for this approach stems from the observation that while ColBERT and its variants have demonstrated strong performance for text retrieval, there is a relative scarcity of similarly scaled-up ColBERT-style models that reach billions of parameters. Consequently, existing smaller ColBERT-style models might not always match the performance of large-scale single-vector embedding models like NV-Embed-2. To approximate this capability, we repurposed the textual encoder component of the ColPali model. The textual encoder of ColPali, based on a PaliGemma variant with 3 billion parameters, effectively serves as a scaled-up ColBERT-style model. This encoder processes the generated captions into multi-vector, per-token embeddings. Textual queries are encoded similarly, and relevance is computed via late-interaction.
Second, to explore late-interaction mechanisms for textual retrieval on captions, we adapted the ColPali architecture in an unconventional manner, termed "Textual ColPali". While ColBERT \cite{10.1145/3397271.3401075} and its variants demonstrate strong text retrieval performance, existing ColBERT-style models are relatively small and may underperform compared to large-scale single-vector embedding models like NV-Embed-2. To address this limitation, we repurposed ColPali's textual encoder component, which effectively serves as a scaled-up ColBERT-style model. This encoder processes generated captions into multi-vector embeddings, with textual queries encoded similarly and relevance computed via late-interaction.

\subsection{Rerankers}

% On certain experimental configurations, we employ reranking models to refine the initial retrieval results. For textual reranking, applied to caption-based retrieval outputs, we evaluated two models:  bge-reranker-v2-gemma\footnote{\url{https://huggingface.co/BAAI/bge-reranker-v2-gemma}} (2.5B parameters) and ms-marco-MiniLM-L-12-v2\footnote{\url{https://huggingface.co/cross-encoder/ms-marco-MiniLM-L-12-v2}} (33M parameters). We deliberately chose to include both a large reranker to ensure maximum retrieval quality, and a smaller, faster reranker for scenarios requiring efficiency and ease of deployment. For configurations involving direct visual retrieval, we utilize multimodal rerankers capable of assessing image-text relevance. Specifically, we use jina-reranker-m0\footnote{https://huggingface.co/jinaai/jina-reranker-v1-m0-base} and lightonai/MonoQwen2-VL-v0.1\footnote{https://huggingface.co/lightonai/MonoQwen2-VL-v0.1}, both of which are vision-language models that determine the relevance between a slide image and a textual query by processing features from both modalities.
On certain experimental configurations, we employ reranking models to refine initial retrieval results. For textual reranking on caption-based outputs, we evaluate \texttt{bge-reranker-v2-gemma} \cite{chen2024bgem3embeddingmultilingualmultifunctionality} \footnote{{\url{https://huggingface.co/BAAI/bge-reranker-v2-gemma}}} (2.5B parameters) and \texttt{ms-marco-MiniLM-L-12-v2} \footnote{\url{https://huggingface.co/cross-encoder/ms-marco-MiniLM-L12-v2}} (33M parameters), selecting both a large model for maximum quality and a smaller model for efficiency. For direct visual retrieval, we use multimodal rerankers \texttt{jina-reranker-m0} \footnote{\url{https://huggingface.co/jinaai/jina-reranker-m0}} and \texttt{MonoQwen2-VL-v0.1} \footnote{\url{https://huggingface.co/lightonai/MonoQwen2-VL-v0.1}}, both VLMs that assess image-text relevance by processing features from both modalities.

\subsection{Experimental Configurations}

Our experimental design systematically evaluates numerous configurations to assess component performance and contributions within slide retrieval pipelines. We start with evaluating simple, single retrieval strategies including BM25 and dense retrieval via semantic similarity on generated slide captions, direct visual retrieval using ColPali on slide images, and text-based late-interaction using Textual ColPali on captions.

We then explore composite retrieval strategies that combine outputs from two distinct methods. For a target list of $k$ slides, we retrieve $k/2$ from each method, handling overlaps by creating unique document lists padded with additional documents from one strategy. We also employ RRF to combine rankings from different retrievers.

Each configuration is evaluated with reranking stages using appropriate reranker models: textual rerankers for caption-based and Textual ColPali methods, visual rerankers for Visual ColPali and composite strategies. When no reranker is employed, we retrieve the top $n=10$ results, whereas configurations utilizing a reranker first retrieve $k=100$ candidates which are then reranked to produce the final top $n=10$ results. Most configurations are tested with all relevant rerankers, though some use only the best-performing reranker for efficiency.

\subsection{Baselines}

% BM25 on OCR text serves as our primary text-based baseline. This traditional information retrieval approach first extracts text from slides using optical character recognition (OCR), then applies the BM25 ranking function to retrieve relevant slides based on lexical matching between the query and the OCR text. While BM25 is known for its effectiveness in text retrieval, its performance on slides may be impacted by OCR errors and the loss of visual context that is crucial for understanding slide content. For OCR text extraction, we applied Tesseract to the SlideVQA dataset and utilized the OCR annotations provided by the authors for the LPM dataset.
To contextualize the performance of the methods investigated in this study, we compare them against several established baselines. The first approach extracts text from slides using OCR, then applies BM25 ranking for lexical matching between queries and OCR text. We applied Tesseract for OCR extraction on SlideVQA and used author-provided OCR annotations for LPM.

% For multimodal baselines, we employ CLIP (Contrastive Language-Image Pre-training) in both zero-shot and fine-tuned configurations. Specifically, we use the \texttt{clip-vit-large-patch14-336} \footnote{\url{https://huggingface.co/openai/clip-vit-large-patch14-336}} variant for our experiments. In the zero-shot setting, we use the pre-trained CLIP model without any task-specific adaptation, leveraging its ability to encode both images and text into a shared vector space. This allows for direct similarity comparison between query embeddings and slide embeddings without requiring any training data from our target domains. The zero-shot CLIP baseline helps us understand the inherent effectiveness of pre-trained multimodal representations for slide retrieval.
For multimodal baselines, we employ CLIP in both zero-shot and fine-tuned configurations using the \texttt{clip-vit-large-patch14-33\-6} \footnote{\url{https://huggingface.co/openai/clip-vit-large-patch14-336}} variant. In the zero-shot setting, we use the pre-trained model without task-specific adaptation.

% --- Third baseline - Option 2 --- 
Our third baseline extends CLIP through task-specific fine-tuning using contrastive learning on our slide datasets. Specifically, for each training instance in a batch of size \(B\), we have one query \(\mathbf{q}_i\), one positive slide embedding \(\mathbf{d}_i^+\), and one negative slide embedding \(\mathbf{d}_i^-\). Additionally, all other negative samples \(\{\mathbf{d}_j^-\}_{j \neq i}\) in the batch act as negative samples for the training instance. We optimize a standard InfoNCE loss with in-batch negatives:
\[
\mathcal{L} = -\frac{1}{B} \sum_{i=1}^{B} \log \Biggl(
    \frac{
        \exp \bigl(\mathrm{sim}(\mathbf{q}_i,\mathbf{d}_i^+)/\tau\bigr)
    }{
        \exp \bigl(\mathrm{sim}(\mathbf{q}_i,\mathbf{d}_i^+)/\tau\bigr)
        + \sum_{j = 1}^{B} \exp \bigl(\mathrm{sim}(\mathbf{q}_i,\mathbf{d}_j^-)/\tau\bigr)
    }
\Biggr),
\]
where \(\mathrm{sim}(\cdot)\) denotes cosine similarity and \(\tau\) is a temperature parameter. 

% For the SlideVQA dataset, we construct our training set triplets utilizing the dataset's QA train split. Each QA sample includes a query and its corresponding evidence slide within the slide deck, which serves as the positive example. To generate negatives for contrastive learning, we randomly sample a slide from the same deck that is not labeled as evidence for the given query. This amounts to 10,617 training instances. For the LPM dataset, we adopted the same methodology as that used for SlideVQA. Each sample consists of a transcript segment paired with its corresponding evidence slide from the slide deck (the complete set of slides from a lecture), serving as the positive example. Negative examples were generated following the exact same process as in SlideVQA. This process yielded 2,938 training instances for the LPM dataset.
For SlideVQA, training triplets were created using its QA train split. Each query and its evidence slide formed a positive example. Negatives were randomly sampled from non-evidence slides of the same deck, yielding 10,617 training instances. For LPM, we apply the same methodology using transcript segments paired with their corresponding evidence slides as positives and random non-evidence slides from the same lecture as negatives, generating 2,938 training instances. We fine-tune CLIP for 5 epochs with a batch size of 6, a learning rate of \(3 \times 10^{-5}\), and \(\tau=0.07\).

\subsection{Evaluation}

To assess our approach, we utilized the SlideVQA and LPM test sets. The SlideVQA test set consists of 2,215 triplets in the form of \(\langle \text{Query}, \text{Answer}, \text{Evidence Slides} \rangle\). Since we are only focusing on the retrieval component, we omit the \(\text{Answer}\). Of these samples, 567 (25.6\%) require evidence from more than one slides, while the remaining can be answered using a single slide. 

The training and test sets from the LPM dataset were originally designed for a different purpose (focusing on subfigures from slides and creating instances for each subfigure). To better align with our objectives, we created a new training and test set from the LPM dataset, using the transcript segments spoken during the display of each slide as queries corresponding to that slide. Our new training set comprises 2,938 instances, while the test set includes 838 instances. 

We evaluated retrieval effectiveness using NDCG@10 and Recall@10 as performance metrics. In addition, we measured the inference time for each retrieval method—with and without the inclusion of a re-ranker—reported in seconds. Storage requirements were also assessed and are reported in gigabytes (GB).

In Figure \ref{fig:dataset_examples_combined}, we present representative examples from both the SlideVQA and LPM datasets to illustrate their distinct characteristics, with example queries shown for both datasets. The examples highlight how SlideVQA questions tend to be focused and specific, while LPM slides are accompanied by longer, more descriptive lecture transcript segments that serve as natural queries.

\subsection{Infrastructure and Implementation Details}

% LLM prompt: Explain that we tried to simulate a low-resource scenario suitable for real-world industry on budget constraints. we used an RTX 3090 24GB for all our experiments. For storing the embeddings and captions, we mostly used an elasticsearch server. we stored the nvidia-embed-v2 embeddings there and made the similarity queries on elastic and the captions. same for bm25. all the colpali embeddings are stored locally in .pt tensor files at fp16 precision (in order to save space, since we load them in VRAM when doing the retrieval) instead of saving them on elasticsearch, since at the time of experimentins elasticsearch did not support late-interaction style retrieval. We run all models in fp32 precision, except of the BGEE reranker,  which we turn down to fp16 when using it with colpali visual in order to allow fit everything in the 24gb vram (we notice that this has minimal drop in perf). all models where used through huggingface and pytorch was used. 

To simulate a low-resource scenario representative of real-world industry budget constraints, all experiments were conducted using a single RTX 3090 GPU with 24GB VRAM. We utilized an Elasticsearch server for indexing both the caption embeddings and the generated slide captions, facilitating efficient similarity queries and BM25 retrieval. Due to the lack of native support for multi-vector late-interaction retrieval in Elasticsearch at the time of experimentation, ColPali’s multi-vector embeddings were instead stored locally as PyTorch tensor files (\texttt{.pt}) using \texttt{fp16} precision to reduce storage footprint and facilitate direct loading into VRAM during retrieval. While all models were generally run at \texttt{fp32} precision, the BGE reranker was specifically operated at \texttt{fp16} precision when used in conjunction with the ColPali visual retrieval pipeline in SlideVQA to ensure the combined memory usage remained within the 24GB VRAM limit. This downcasting was observed to result in a minimal drop in performance. Additionally, the \texttt{jina-reranker-m0} reranker is configured by its creators to load in \texttt{fp16} precision by default. All models were loaded via the Hugging Face repository, with PyTorch serving as the underlying deep learning framework.

\setlength{\tabcolsep}{2pt} % Default is 6pt, reduce as needed
\begin{table*}[!t]
    \small % or \footnotesize
    \centering
    \caption{Ablation on retrievers, captioning models, and rerankers for SlideVQA and LPM dataset. For brevity, the captioning models \texttt{Molmo-7B-D-0924} and \texttt{Gemma3-27B-IT-QAT} are referred to as "Molmo" and "Gemma3" respectively. Similarly, rerankers \texttt{ms-marco-MiniLM-L-12-v2}, \texttt{bge-reranker-v2-gemma}, \texttt{MonoQwen2-VL-v0.1}, and \texttt{jina-reranker-m0}" are denoted as "MiniLM", "BGE", "MonoQwen2", and "Jina" respectively. "Inf. Time" measures the average inference time in seconds per query. "Storage" is measured in GB for the entire dataset. For configurations with rerankers, "Inf. Time" shows the sum of initial retrieval time and reranking time (e.g., 0.22 + 0.09 means 0.22s for retrieval and 0.09s for reranking).} % Adjusted caption slightly
    \begin{tabular}{ccccccccccc} % Changed from ccccccc to ccccccccccc (3 + 4 + 4 = 11 columns)
        \toprule
        \multirow{2}{*}{Retrieval method} & \multirow{2}{*}{Captioning model} & \multirow{2}{*}{Reranker} & \multicolumn{4}{c}{SlideVQA} & \multicolumn{4}{c}{LPM} \\
        \cmidrule(lr){4-7} \cmidrule(lr){8-11} % Adjusted cmidrule indices
         & & & NDCG@10 & Recall@10 & Inf. Time & Storage & NDCG@10 & Recall@10 & Inf. Time & Storage \\ % Added new headers

        \midrule
        \multirow{1}{*}{BM25 on OCR} & \multirow{1}{*}{-} & - & 54.3 & 61.6 & 0.01 & 0.01 & 55.9 & 65.1 & 0.01 & 0.003 \\

        \midrule
        \multirow{1}{*}{CLIP (zero-shot)} & \multirow{1}{*}{-} & - & 45.3 & 58.4 & 0.34 & 0.15 & 45.4 & 62.1 & 0.06 & 0.02 \\

        \midrule
        \multirow{1}{*}{CLIP (fine-tuned)} & \multirow{1}{*}{-} & - & 49.5 & 62.6 & 0.32 & 0.15 & 53.2 & 69.8 & 0.05 & 0.02 \\

        \midrule
        \multirow{1}{*}{DSE (zero-shot)} & \multirow{1}{*}{-} & - & 64.0 & 76.1 & 0.04 & 0.11 & 63.4 & 76.0 & 0.04 & 0.11 \\

        \midrule
        \multirow{1}{*}{DSE (SlideVQA)} & \multirow{1}{*}{-} & - & 75.3 & 84.6 & 0.04 & 0.11 & - & - & - & - \\
        
        \midrule
        \multirow{3}{*}{BM25} & \multirow{3}{*}{Molmo} & - & 63.6 & 73.0 & 0.04 & \multirow{3}{*}{0.04} & 57.9 & 69.8 & 0.05 & \multirow{3}{*}{0.006} \\
         & & MiniLM & 72.6 & 80.8 & 0.22 + 0.09 &  & 51.3 & 65.8 & 0.24 + 0.17 &  \\
         & & BGE & 75.9 & 82.1 & 0.23 + 7.72 &  & 68.1 & 79.4 & 0.24 + 12.41 &  \\
        \midrule
        \multirow{3}{*}{Neural} & \multirow{3}{*}{Molmo} & - & 65.3 & 76.4 & 0.05 & \multirow{3}{*}{4.23} & 67.0 & 81.0 & 0.04 & \multirow{3}{*}{0.72} \\
         & & MiniLM & 72.4 & 81.3 & 0.24 + 0.09 &  & 50.7 & 66.3 & 0.23 + 0.17 &  \\
         & & BGE & 76.2 & 83.2 & 0.25 + 7.42 &  & 70.4 & 84.0 & 0.24 + 12.41 &  \\
        \midrule
        \multirow{3}{*}{BM25 + Neural} & \multirow{3}{*}{Molmo} & - & 68.9 & 81.6 & 0.10 & \multirow{3}{*}{4.28} & 66.5 & 79.4 & 0.09 & \multirow{3}{*}{0.75} \\
         & & MiniLM & 73.9 & 83.4 & 0.49 + 0.09 &  & 50.9 & 66.7 & 0.37 + 0.17 & \\
         & & BGE & 78.3 & 85.9 & 0.48 + 7.39 &  & 70.4 & 83.6 & 0.38 + 12.40 & \\

        \midrule
        \multirow{3}{*}{BM25} & \multirow{3}{*}{Gemma3} & - & 68.8 & 78.5 & 0.04 & \multirow{3}{*}{0.04} & 57.1 & 68.3 & 0.05 & \multirow{3}{*}{0.006} \\
         & & MiniLM & 76.0 & 84.6 & 0.19 + 0.23 &  & 48.3 & 63.9 & 0.20 + 0.23 &  \\
         & & BGE & 82.2 & 87.5 & 0.20 + 7.41 &  & 67.0 & 77.8 & 0.21 + 13.81 & \\
        \midrule
        \multirow{3}{*}{Neural} & \multirow{3}{*}{Gemma3} & - & 63.5 & 76.4 & 0.04 & \multirow{3}{*}{4.21} & 63.9 & 79.4 & 0.03 & \multirow{3}{*}{0.73} \\
         & & MiniLM & 75.1 & 83.9 & 0.20 + 0.22 &  & 44.9 & 62.4 & 0.19 + 0.23 & \\
         & & BGE & 81.4 & 87.4 & 0.21 + 7.34 &  & 69.5 & 82.3 & 0.21 + 12.94 & \\
        \midrule
        \multirow{3}{*}{BM25 + Neural} & \multirow{3}{*}{Gemma3} & - & 66.5 & 84.3 & 0.08 & \multirow{3}{*}{4.29} & 63.8 & 79.1 & 0.08 & \multirow{3}{*}{0.78} \\
         & & MiniLM & 76.6 & 86.1 & 0.30 + 0.22 &  & 46.0 & 63.2 & 0.32 + 0.23 &  \\
         & & BGE & 83.9 & 90.6 & 0.31 + 7.45 &  & 69.6 & 82.3 & 0.33 + 13.34 & \\

        \midrule
        \multirow{3}{*}{ColPali (Visual)} & \multirow{3}{*}{-} & - & 82.7 & 89.9 & 0.21 & \multirow{3}{*}{12.90} & 66.7 & 80.6 & 0.04 & \multirow{3}{*}{2.22} \\
         & & MonoQwen2 & 84.3 & 90.9 & 0.22 + 14.40 &  & 66.9 & 82.3 & 0.04 + 18.45 &  \\
         & & Jina & 86.9 & 92.9 & 0.22 + 14.24 &  & 73.6 & 84.8 & 0.04 + 11.26 &  \\

        \midrule
        \multirow{3}{*}{ColPali (Textual)} & \multirow{3}{*}{Molmo} & - & 76.1 & 84.7 & 0.22 & \multirow{3}{*}{2.38} & 65.2 & 78.1 & 0.03 & \multirow{3}{*}{0.41} \\
         & & MiniLM & 74.8 & 84.8 & 0.22 + 0.13 &  & 46.1 & 63.0 & 0.03 + 0.17 & \\
         & & BGE & 79.2 & 87.3 & 0.22 + 8.01 &  & 69.3 & 81.8 & 0.04 + 11.87 & \\

        \midrule
        \multirow{3}{*}{ColPali (Textual)} & \multirow{3}{*}{Gemma3} & - & 78.4 & 86.6 & 0.22 & \multirow{3}{*}{2.87} & 64.9 & 77.6 & 0.03 & \multirow{3}{*}{0.45} \\
         & & MiniLM & 77.4 & 87.0 & 0.22 + 0.20 &  & 52.0 & 67.3 & 0.03 + 0.23 & \\
         & & BGE & 84.0 & 90.5 & 0.23 + 7.25 &  & 70.1 & 82.5 & 0.04 + 12.29 & \\

        \midrule
        \multirow{3}{*}{\shortstack{ColPali (Visual) + \\ ColPali (Textual)}} & \multirow{3}{*}{Gemma3} & - & 83.8 & 91.1 & 0.44 & \multirow{3}{*}{15.77} & 64.5 & 74.3 & 0.08 & \multirow{3}{*}{2.67} \\
         & & BGE & 85.0 & 91.9 & 0.46 + 3.66 & & 67.8 & 78.1 & 0.08 + 14.06 & \\
         & & Jina & 86.9 & 92.7 & 0.45 + 13.94 & & 72.2 & 83.0 & 0.08 + 10.95 & \\

        \midrule
        \multirow{3}{*}{\shortstack{BM25 + \\ ColPali (Textual)}} & \multirow{3}{*}{Gemma3} & - & 75.9 & 87.5 & 0.25 & \multirow{3}{*}{3.30} & 61.3 & 76.6 & 0.07 & \multirow{3}{*}{0.46} \\
         & & BGE & 84.5 & 91.1 & 0.39 + 7.44 & & 69.6 & 82.2 & 0.16 + 13.67 & \\
         & & Jina & 86.1 & 91.6 & 0.34 + 14.15 & & 72.6 & 83.5 & 0.15 + 11.31 & \\

        \midrule
        \multirow{3}{*}{\shortstack{RRF[ColPali (Visual), \\ ColPali (Textual)]}} & \multirow{3}{*}{Gemma3} & - & 81.0 & 89.3 & 0.45 & \multirow{3}{*}{15.77} & 66.8 & 80.7 & 0.08 & \multirow{3}{*}{2.67} \\
         & & BGE & 84.9 & 91.8 & 0.45 + 3.68 &  & 69.9 & 82.8 & 0.08 + 4.9 & \\
         & & Jina & \textbf{87.0} & \textbf{92.9} & 0.47 + 12.03 & & \textbf{73.6} & \textbf{84.9} & 0.09 + 11.44 & \\
        
        \bottomrule
    \end{tabular}
    \label{tab:retriever_reranker_performance}
\end{table*}

\section{Results}

Table \ref{tab:retriever_reranker_performance} presents the results for all evaluated retrieval configurations, including performance metrics, inference times, and storage demands on the SlideVQA and LPM datasets.

\subsection{Baselines}

The traditional BM25 approach applied to OCR-extracted text achi\-eves moderate performance with NDCG@10 scores of 54.3\% and 55.9\% on SlideVQA and LPM respectively. Among the multimodal baselines, zero-shot CLIP underperforms BM25 on both datasets, achieving only 45.3\% NDCG@10 on SlideVQA. Fine-tuning CLIP on task-specific data provides modest improvements, reaching 49.5\% on SlideVQA and 53.2\% on LPM, though still falling short of the BM25 baseline. In contrast, DSE demonstrates substantially stronger zero-shot performance at 64.0\% NDCG@10 on SlideVQA, with its fine-tuned variant on SlideVQA achieving 75.3\% on the same dataset, validating the effectiveness of VLMs specifically tuned for document understanding.

\subsection{Sparse and Bi-Encoder Retrieval on Slide Captions}

Caption-based retrieval strategies demonstrate notable efficacy, often outperforming baseline methods across both datasets. The choice of VLM for generating these captions plays an important role. Gemma3-27B-IT captions generally yield superior results on SlideVQA, particularly when enhanced by reranking. For instance, a hybrid approach combining BM25 and neural retrieval (using NV-Embed-2 embeddings) on Gemma3 captions, followed by BGE reranking, achieved the highest NDCG@10 of 83.9\% and Recall@10 of 90.6\% on SlideVQA among these experiments. This significantly surpasses the fine-tuned DSE baseline (75.3\% NDCG@10). Interestingly, on the LPM dataset, Molmo-7B-D-0924 captions led to better performance, with neural retrieval on Molmo captions coupled with BGE reranking achieving 70.4\% NDCG@10 and 84.0\% Recall@10, slightly edging out its hybrid counterpart and configurations using Gemma3 captions. The hybrid BM25+Neural approach demonstrates clear advantages over either method in isolation on SlideVQA, achieving up to 68.9 NDCG@10 without reranking, confirming that lexical and semantic signals can provide complementary relevance information.  

The application of rerankers substantially influences performance, though with practical trade-offs. For example, on SlideVQA, it boosted the hybrid Gemma3-caption method from 66.5\% to 83.9\% NDCG@10. However, this gain incurs a considerable latency penalty, with BGE adding approximately 7.4 seconds per query on SlideVQA and over 12 seconds on LPM, potentially limiting its use in real-time RAG applications. Conversely, the lightweight MiniLM reranker, while offering a speed advantage (0.09s-0.23s), provided modest gains on SlideVQA and detrimentally affected performance on the LPM dataset. In terms of space utilization, caption-based methods are highly efficient for storing the captions themselves (e.g., 0.04 GB for SlideVQA with Molmo). While neural retrieval on these captions introduces larger storage needs due to dense embeddings (e.g., 4.21 GB for NV-Embed-2 with Gemma3 captions on SlideVQA), this is still considerably less than visual late-interaction models like ColPali (Visual) (12.90 GB). A simple BM25 approach on captions offers an extremely lightweight alternative in both storage and inference time (0.04-0.05s without reranking), providing a strong, efficient baseline.

\subsection{Late-Interaction Retrieval}

Direct visual retrieval using ColPali (Visual) demonstrates strong performance, particularly when paired with a visual reranker. On its own, ColPali (Visual) achieves 82.7\% NDCG@10 on SlideVQA and 66.7\% on LPM. The addition of the Jina visual reranker elevates these scores to a leading 86.9\% NDCG@10 on SlideVQA and 73.6\% on LPM, establishing this combination as the top-performing single-retriever method. However, this peak performance comes at a significant cost. ColPali (Visual) requires considerable storage (12.90 GB for SlideVQA image embeddings) and the Jina reranking step adds substantial latency (approximately 14.24 seconds per query on SlideVQA and 11.26 seconds on LPM), making it resource-intensive for applications demanding rapid responses. The MonoQwen2 reranker also boosts ColPali (Visual) but to a lesser extent than Jina.

The textual adaptation, "Textual ColPali," applied to slide captions, offers a compelling alternative with a significantly smaller storage footprint. Using Gemma3-generated captions, Textual ColPali achieves 78.4\% NDCG@10 on SlideVQA (2.87 GB storage) and 64.9\% on LPM (0.45 GB storage) without reranking. When combined with the BGE textual reranker, performance on SlideVQA with Gemma3 captions rises to 84.0\% NDCG@10. While the BGE reranker still introduces notable latency (around 7.25 seconds on SlideVQA with Textual ColPali on Gemma3 captions), this configuration is highly competitive with the best caption-based methods (like hybrid BM25+Neural with BGE at 83.9\% NDCG@10) and is more storage-efficient than ColPali (Visual). Performance with Molmo captions for Textual ColPali is generally lower than with Gemma3 on SlideVQA but slightly better on LPM before reranking.

Hybridizing ColPali (Visual) and ColPali (Textual) (using Gemma 3 captions) through methods like simple combination or RRF further refines performance. The RRF combination, when reranked with Jina, achieves an impressive 87.0\% NDCG@10 on SlideVQA and 73.6\% on LPM, matching or slightly exceeding the ColPali (Visual) + Jina setup. Similarly, combining BM25 (on Gemma3 captions) with Textual ColPali (Gemma3 captions) and reranking with Jina yields strong results (86.1\% NDCG@10 on SlideVQA), outperforming standalone Textual ColPali + BGE and demonstrating the benefits of integrating sparse signals with late-interaction textual models.

\subsection{Discussion}

\subsubsection{Key Findings}

Our findings demonstrate that hybrid textual retrieval methods applied to VLM-generated slide captions (e.g., BM25 combined with neural embeddings and a BGE reranker) can significantly outperform direct multimodal encoding approaches like fine-tuned DSE. However, the highest retrieval efficacy was achieved by the late-interaction ColPali (Visual) model when augmented with a Jina visual reranker, or by fusing ColPali (Visual) with its textual counterpart on captions via RRF. Textual ColPali itself, applied to captions, offers a competitive and notably more storage-efficient alternative. Across all high-performing configurations, powerful rerankers proved critical for achieving top scores but invariably introduced substantial latency, underscoring a core trade-off with storage requirements and computational cost. The choice of VLM for captioning (Molmo vs. Gemma3) also revealed dataset-dependent performance variations.

\subsubsection{What's the Best Way to Retrieve Slides?}

% The "best way" to retrieve slides is nuanced and contingent upon specific application constraints. For scenarios prioritizing maximal retrieval accuracy where computational resources and latency are less critical, ColPali (Visual) with a Jina reranker or its RRF hybrid with Textual ColPali stands out as optimal. However, for RAG systems that demand a more practical balance between high accuracy, lower latency, and manageable storage, hybrid textual retrieval on high-quality captions (such as BM25+Neural+BGE) or Textual ColPali with a BGE reranker are highly effective. These results strongly suggest that transforming the multimodal slide retrieval problem into the textual domain via rich VLM-generated captions allows leveraging the strengths of mature and powerful textual IR techniques, often yielding a better practical solution than direct multimodal encoding. For highly resource-constrained environments, even a simple BM25 on good quality captions provides a respectable and efficient baseline. The significant latency introduced by the most effective rerankers (both visual and textual) remains a key challenge for deploying these top-performing configurations in interactive RAG systems, emphasizing the ongoing need for retrieval strategies that are simultaneously accurate, fast, and resource-efficient.
The optimal slide retrieval method depends on application constraints. For maximal accuracy, if resources and latency are secondary, ColPali (Visual) with a Jina reranker or its RRF hybrid with Textual ColPali is optimal. However, for RAG systems needing a balance of high accuracy, low latency, and manageable storage, hybrid textual retrieval on high-quality captions (e.g., BM25+Neural+BGE) or Textual ColPali with a BGE reranker are highly effective. These findings suggest that converting the multimodal slide retrieval problem to a textual one allows leveraging mature textual IR techniques, often yielding a more practical solution than direct multimodal encoding. In highly resource-constrained environments, even simple BM25 on quality captions provides a respectable and efficient baseline. The significant latency from top-performing rerankers (both visual and textual) remains a key challenge for their deployment in interactive RAG systems.

\section{Conclusion and Future Work}

This work provides a detailed comparison of slide retrieval techniques, revealing that while SOTA accuracy often involves resource-intensive late-interaction models and powerful rerankers, transforming slides into rich textual captions enables strong performance with more practical resource profiles. The presented analysis of retrieval efficacy against latency and storage offers crucial insights for industry practitioners aiming to deploy effective and efficient slide retrieval systems, guiding the selection of optimal strategies based on specific real-world application constraints and performance targets in corporate and academic settings.

Future efforts will focus on fine-tuning ColPali on slide-specific data to further enhance its leading performance. We also aim to develop more advanced textual late-interaction models, improving the efficacy of storage-efficient caption-based retrieval. Critically, research into faster and potentially domain-adapted reranking models is vital to reduce the substantial latency introduced by current top-performing rerankers, facilitating their practical use in interactive slide retrieval systems.

\section{Acknowledgments}
This research work was funded by JOHNSON AND
JOHNSON SERVICES, INC.

\section{GenAI Usage Disclosure}
Generative AI tools were utilized during the preparation of this manuscript for editing, improving the quality of existing text, and generating minor textual elements.

\bibliographystyle{ACM-Reference-Format}
\bibliography{citations}

\end{document}